\documentclass[11pt]{article}

\usepackage{microtype} 
\usepackage{booktabs}  
\usepackage{url}  
\usepackage{graphicx} 
\usepackage{float}
\usepackage{xcolor}
\usepackage{amsmath}
\usepackage{amsthm}
\usepackage{times}
\usepackage{soul}
\usepackage{url}
\usepackage[utf8]{inputenc}
\usepackage[small]{caption}
\usepackage{afterpage}
\usepackage{algorithm}
\usepackage{algorithmic}
\usepackage{graphicx}
\usepackage{amssymb}
\usepackage{pifont}
\newcommand{\cmark}{\ding{51}}%
\newcommand{\xmark}{\ding{55}}%
\usepackage{longtable}

\usepackage{wrapfig}
\usepackage{subcaption}
\usepackage[final]{automl}
\usepackage{natbib}
\title{CLAMS: A System for Zero-Shot Model Selection for Clustering}

\author[1]{\nameemail{Prabhant Singh}{p.singh@tue.nl}}
\author[1]{\nameemail{Pieter Gijsbers}{p.gijsbers}}
\author[1]{\nameemail{Murat Onur Yildirim}{m.o.yildirim@tue.nl}}
\author[1]{\nameemail{Elif Ceren Gok}{ e.c.gok@tue.nl}}
\author[1]{\nameemail{Joaquin Vanschoren}{j.vanschoren@tue.nl}}

\affil[1]{Eindhoven University of Technology}

\hypersetup{%
  pdfauthor={}, 
  pdftitle={},
  pdfsubject={},
  pdfkeywords={}
}

\begin{document}

\maketitle

\begin{abstract}
We propose an AutoML system that enables model selection on clustering problems by leveraging optimal transport-based dataset similarity. Our objective is to establish a comprehensive AutoML pipeline for clustering problems and provide recommendations for selecting the most suitable algorithms, thus opening up a new area of AutoML beyond the traditional supervised learning settings. We compare our results against multiple clustering baselines and find that it outperforms all of them, hence demonstrating the utility of similarity-based automated model selection for solving clustering applications.
\end{abstract}
\section{Introduction}
Clustering plays a significant role in data analysis. It can be useful for exploratory data analysis, user profiling, and medical analysis tasks~\citep{clusteringapp}. In the last few decades, many clustering methods and tools have been developed, each with their own strengths and weaknesses, making it difficult to select the best ones. For instance, the scikit-learn library~\citep{scikit-learn} contains 13 clustering algorithms, each with several hyperparameters to tune. This makes it very difficult for non-experts to select the best models and hyperparameters, further exacerbated by the fact that there is no golden metric to optimize for.

Indeed, clustering by its very nature is unsupervised. The lack of a ground truth makes it difficult to use traditional AutoML techniques, as they often rely on known labels to evaluate candidate solutions. Previous automated clustering approaches evaluate clustering methods via multiple internal metrics (which do not require labels), such as Calinkski-Harbarnz or the Silhouette score, or external metrics (which require labels) such as adjusted mutual information (AMI) and rand index. These metrics are also called Cluster Validity Indices (CVIs). One can optimize internal CVIs and aim to achieve optimal performance over external CVIs, but to the best of our knowledge, there has not been any significant work proving a strong correlation between internal and external CVIs. We cannot optimize for an external CVI for a dataset where no labels are available, which makes model selection for a new dataset without labels an extremely hard problem.

\par
In this work, we propose an entirely unsupervised, zero-shot model selection framework that automates clustering by `pretraining' on previous tasks with internal and external CVIs. As such, it takes a significant step forward to enabling AutoML for clustering in real-world settings. We make the following contributions:
\begin{enumerate}
    \item CLAMS (Clustering with Automated Machine Learning System) is a standalone, open-source AutoML tool for automated clustering. 
    \item CLAMS-OT: A general zero-shot model recommendation system leveraging dataset similarity based on optimal transport (OT).
\end{enumerate}
\par
 We evaluate our system on a wide range of clustering problems and find that it significantly outperforms all baselines, as demonstrated by Bayesian-Wilcoxon signed-rank tests~\citep{rope2, Ropetutorial} and critical difference diagrams~\citep{demsar2006statistical}. We provide all code and benchmarking problems so that these results can be easily reproduced.

\section{Background}

\subsection{AutoML for Clustering}
Automated machine learning provides methods and tools to make machine learning solutions available to people without the time or ML expertise to finetune ML models. Some examples of AutoML systems include Auto-Sklearn~\citep{feurer-arxiv20a}, GAMA~\citep{gama}, and FLAML~\citep{Wang2021FLAMLAF}. These systems enable the automation of model selection and hyperparameter optimization (HPO) for supervised classification and regression tasks. However, very little research in current AutoML literature addresses automated clustering. 

Earlier studies focused on optimizing the number of clusters~\citep{tseng2001genetic, liu2011automatic, das2007automatic, saha2013generalized}. Later work incorporates algorithm selection, sometimes through meta-learning, or hyperparameter optimization beyond just selecting the number of clusters. Table~\ref{tab:overview} provides an overview of the different methods. Most of them optimize a CVI metric or apply meta-learning techniques based on dataset similarity measured via meta-features such as Landmarkers, Clustering-oriented Meta-feature Extraction (CME), and Cluster Cardinality Estimation (CCE).

\begin{table}[h]
\small
\begin{tabular}{ccccc}
\hline
                           & \textbf{Meta-Learning}     & \textbf{Algorithm Selection}   & \textbf{HPO} & \textbf{Unsupervised}\\ \hline
\cite{tseng2001genetic}           & \xmark           & \xmark      & \cmark & \xmark \\
\cite{liu2011automatic}           & \xmark           & \xmark      & \cmark & \xmark \\
\cite{das2007automatic}           & \xmark           & \xmark      & \cmark & \xmark \\
\cite{saha2013generalized}        & \xmark           & \xmark      & \cmark & \xmark \\
\cite{saez2019meta}               & CCE              & \cmark      & \xmark  & \xmark \\
\cite{pimentel2019new}            & CME, CVI          & \cmark      & \xmark  & \xmark \\
\cite{de2008ranking}              & Landmarkers       & \cmark      & \xmark  & \xmark \\
\cite{soares2009analysis}         & Landmarkers       & \cmark      & \xmark  & \xmark \\
\cite{tschechlov2021automl4clust} & \xmark            & \cmark      & \cmark & \xmark \\
\cite{poulakis2020autoclust}      &  CVI             & \cmark      & \cmark & \xmark \\
\cite{liu2021autocluster}         &  CME, CVI         & \cmark      & \cmark & \cmark \\
Ours                               &   OT              & \cmark      & \cmark & \cmark \\ \hline
\end{tabular}
\caption{An overview of earlier work in automated clustering showing which parts of automated model selection and hyperparameter optimization are employed in each.}
\label{tab:overview}
\end{table}

\subsection{Meta-learning}
Meta-learning (or \textit{learning to learn}) is often employed in AutoML to learn from the historical performance of machine learning models on a variety of tasks and use this knowledge to find better models for new tasks~\citep{Vanschoren2018MetaLearningAS}. It can help to speed up the model selection process and find better architectures. Meta-learning is often used to \textit{warm-start} the search for optimal models by choosing good initial hyperparameters or reducing the search space that AutoML algorithms have to explore. Multiple strategies have been proposed for meta-learning in AutoML including transferring priors in Bayesian optimization, K-nearest datasets~\citep{NIPS2015_11d0e628}, portfolio selection~\citep{feurer-arxiv20a}, and optimal transport~\citep{metabu}.

In clustering, few studies have considered meta-learning for algorithm selection without hyperparameter optimization~\citep{saez2019meta, pimentel2019new}. \cite{de2008ranking, soares2009analysis} require ground truth labels since they used external metrics during optimization. AutoML4Clust~\citep{tschechlov2021automl4clust} and AutoClust ~\citep{poulakis2020autoclust} leverage Bayesian Optimization to select the best clustering algorithm and hyperparameters for a given dataset, but they do not support pipelines with data preprocessing steps. AutoCluster~\citep{autocluster} uses meta-learning for creating ensembles of multiple clustering algorithms.\footnote{We tried to add autocluster, but encountered errors we could not resolve ourselves. We could not reach the authors.} ISAC~\citep{ISAC} is a similar work methodologically from the field of algorithmic configuration. In this work, we focus on optimal transport based meta-learning as that can be used in unsupervised tasks. In the next section, we give a short overview of optimal transport and dataset similarity. 

\subsection{Optimal Transport}
Optimal transport (OT) or transportation theory, also known as Kantorovich–Rubinstein duality, is a problem that deals with the transportation of masses from source to target. This problem is also called the Monge–Kantorovich transportation problem~\citep{Villani2008OptimalTO}. In recent years, OT has gained significant attention from the machine learning community, as it provides a powerful framework for designing algorithms that can learn to match two probability distributions, which is a common task in image and natural language processing. In this section, we give an introduction to OT and distance measures related to our work. 

In OT, the objective is to minimize the cost of transportation between two probability distributions. For a cost function between pairs of points, we calculate the cost matrix $C$ with dimensionality $n \times m $. The OT problem minimizes the loss function $L_c(P):=\langle C, P\rangle$ with respect to a coupling matrix $P$. A practical and computationally more efficient approach is based on regularization and minimizes $L_c^\epsilon(P):=\langle C, P\rangle + \epsilon \cdot r(P)$ where $r$ is the negative entropy, computed by the Sinkhorn algorithm~\citep{Cuturi2013SinkhornDL}, and $\epsilon$ is a hyperparameter controlling the amount of regularization. A discrete OT problem can be defined with two finite point clouds, 
$\{x^{(i)}\}^{n}_{i=1}$ ,$\{y^{(j)}\}^{m}_{j=1}, x^{(i)},y^{(j)}\in \mathbb{R}^d $, which can be
described as two empirical distributions: $\mu:=\sum^n_{i=1}a_i\delta_{x^{(i)}}, \nu:=\sum^m_{j=1}b_j\delta_{y^{(j)}}$. 
Here, $a$ and $b$ are probability vectors of size $n$ and $m$, respectively, and the $\delta$ is the Dirac delta. 

\subsubsection{Gromov Wasserstein Distance}
In this work, we are interested in the Gromov Wasserstein (GW) distance between these two discrete probability distributions. Gromov Wasserstein allows us to match points within different metric spaces, which is the case in AutoML problems where we have datasets of different dimensionalities. This problem can be written as a function of $(a,A), (b,B)$ between our distributions $A$ and $B$~\citep{Villani2008OptimalTO,gwlr}: 
\begin{equation}
        \text{GW}((a,A),(b,B)) =
        \min_{P\in \Pi_{a,b}} \mathcal{Q}_{A,B}(P)
\end{equation}

where $\Pi_{a,b}:=\{ P \in \mathbb{R}^{n \times m}_+| P\mathbf{1}_m = a, P^{T}\mathbf{1}_n=b\}$ is the set of all possible mappings of points from $A$ to $B$ and the \textit{energy} $\mathcal{Q}_{A,B}$ is a quadratic function of $P$  which can be described as 
\begin{equation}
    \mathcal{Q}_{A,B}(P):= \sum_{i,j,i^{'},j^{'}}(A_{i,i'}-B_{j,j'})^2P_{i,j}P_{i',j'}
\end{equation} 
In this work we are interested in the Entropic Gromov Wasserstein cost~\citep{pmlr-v48-peyre16}:

\begin{equation}\label{eq 2-1}
    \text{GW}_\varepsilon((a,A),(b,B)) = \min_{P\in \Pi_{a,b}} \mathcal{Q}_{A,B}(P) - \varepsilon \cdot H(P)
\end{equation}

where $GW_\epsilon$ is the Entropic Gromov Wasserstein cost between our distributions $A$ and $B$,
$H(P)$ is the Shannon entropy, and $\varepsilon$ a regularization constant. The problem with Gromov Wasserstein is that it is NP-hard and the entropic approximation of GW still has cubic complexity. For practical speedup, we use the Low-Rank Gromov Wasserstein (GW-LR) approximation~\citep{pmlr-v139-scetbon21a, Scetbon2022LowrankOT,gwlr}, which reduces the computational cost from cubic to linear time. \cite{gwlr} consider the Gromov Wasserstein problem with low-rank couplings, linked by a common marginal $g$. Therefore, the set of possible transport plans is restricted to those adopting the factorization of the form $P_r = Q_{diag}(1/g)R^T$, where $Q$ and $R$ are thin matrices with the dimensionality of $n\times r$ and $r\times m$, respectively, and $g$ is an $r$-dimensional probability vector. 

The GW-LR distance is then described as:

\begin{equation}\label{eq 2-2}
\begin{aligned}
    \text{GW-LR}^{(r)} & ((a,A), (b,B)) := \\
                       & \min_{(Q,R,g)\in \mathcal{C}_{a,b,r}}\mathcal{Q}_{A,B}(Q_{diag}(1/g)R^T)
\end{aligned}
\end{equation}

 \section{Problem statement}
We first introduce some notation: 
\begin{itemize}
    \item $\mathcal{D}_{meta}$: A set of $n$ datasets $\{D_1, \cdots, D_n\}$, these are prior datasets with known labels.
    \item $\boldsymbol{A}$: Space of all possible algorithms applicable to the specific machine learning problem. Each algorithm $A \in \boldsymbol{A}$ has a hyperparameter configuration space $\Lambda_A$. An algorithm initialized with hyperparameter configuration $\lambda \in \Lambda_A$ is referred to as $A_{\lambda}$.
    \item $\mathcal{A}$: A set $\{A^*_{\lambda^*_1}, ..., A^*_{\lambda^*_n}\}$ with the best algorithm configuration $A^*_{\lambda^*_i}$ found for each dataset $D_i \in \mathcal{D}_{meta}$ according to a loss function $L$.
\end{itemize}

\textbf{Problem Statement:} Given a new dataset without any labels, our meta-learner needs to select an optimal algorithm with associated hyperparameter configuration from a collection of previously evaluated algorithm configurations. Since we cannot further optimize the given model on the new dataset this is a \textit{zero-shot model recommendation problem}, unless some (downstream) evaluation metric is available.

Formally, given a new \textit{unlabeled} dataset $D_{new}$, select an algorithm with configuration $A^*_{\lambda^*} \in \mathcal{A}$ to employ on $D_{new}$, where $A^*_{\lambda^*_i} $ is the optimal model with tuned hyperparameters for the \textit{labeled} dataset $D_i$ that is most similar to $D_{new}$.
\par
\textbf{Problem Formulation:} For supervised tasks, this problem can be represented as a Combined Algorithm Selection and Hyperparameter optimization (CASH) problem~\citep{Thornton2013}, shown in equation \ref{eq:1}, where $A^*_{\lambda^*}$ is the combination of the optimal learning algorithm from search space $A$ with associated hyperparameter space $\Lambda_A$ evaluated over $k$ cross-validation folds of dataset $D=\{X,y\}$ with training and validation splits for loss measure $L$.
\begin{equation}
\begin{split}
A^*_{\lambda^*} =
\operatorname*{argmin}_{%
       \substack{%
         A \in \boldsymbol{A} \\
         \lambda \in \Lambda_A
       }
     }
\frac{1}{k} 
\sum_{f=1}^{k} L \left( A_\lambda, \big\{\boldsymbol{X}^{train}_f, \boldsymbol{y}^{train}_f\big\}, \big\{\boldsymbol{X}^{val}_f, \boldsymbol{y}^{val}_f\big\} \right)
\end{split}
\label{eq:1}
\end{equation}
\par
The CASH problem from Equation \ref{eq:1} relies on the validation split to optimize for the optimal configuration, because labels are used during training. However, since unsupervised learning algorithms do not use labels while training, validation splits are not relevant and we can instead train on all unlabeled data and use the ground truth labels only for evaluation. Our modified CASH formulation to select the optimal unsupervised algorithm \textbf{with access to labels} during optimization is as follows:

\begin{equation}
\begin{split}
A^*_{\lambda^*} =
\operatorname*{argmin}_{%
       \substack{%
         A \in \boldsymbol{A} \\
         \lambda \in \Lambda_A
       }
     }
 L \left( A_\lambda, \big\{\boldsymbol{X}\}, \big\{\boldsymbol{y}\big\} \right)
\end{split}
\label{eq: 2}
\end{equation}
\par

Our method can also be used with internal metrics $L_{internal}$, which do not require labels, in which case only unlabeled data is needed for optimization:
\begin{equation}
\begin{split}
A^*_{\lambda^*} =
\operatorname*{argmin}_{%
       \substack{%
         A \in \boldsymbol{A} \\
         \lambda \in \boldsymbol{\Lambda_A}
       }
     }
 L_{internal} \left( A_\lambda, \big\{\boldsymbol{X}\} \right)
\end{split}
\label{eq: 3}
\end{equation}
\par
\section{Proposed Framework}
In this section, we describe the components of our framework. Our proposed framework has two components.
\begin{enumerate}
    \item We first propose \textbf{CLAMS}, Clustering with Automated Machine Learning System. CLAMS is a standalone AutoML tool for clustering with any metric. In this work, CLAMS is used for populating the meta-data.
    \item Secondly we propose a dataset similarity indicator with optimal transport distances, which is used for zero-shot model-recommendation 
\end{enumerate}

\subsection{CLAMS: Clustering with Automated Machine Learning System}

\begin{figure}[t]
  \centering
  \includegraphics[width=1\textwidth]{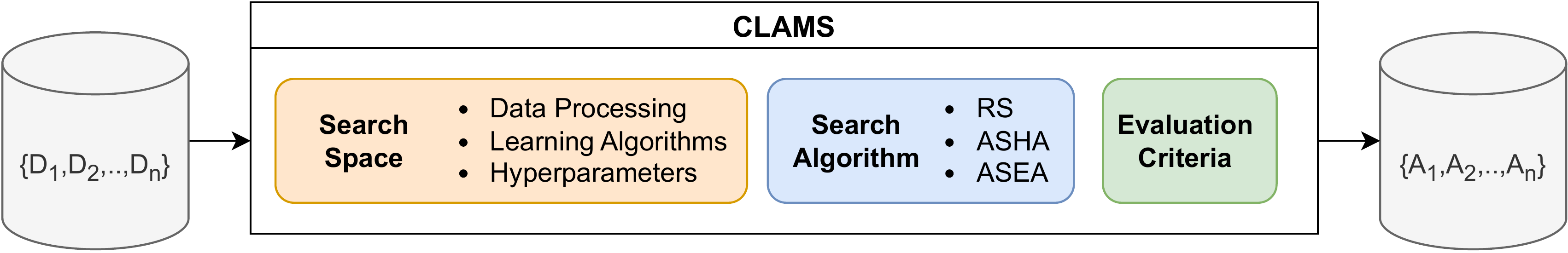}
  \caption{CLAMS Overview}
\end{figure}

In this work, we have devised an Automated Machine Learning framework that we called CLAMS. Our proposed framework comprises a diversified and well-defined search space that encompasses various preprocessing steps and learning algorithms, as well as optimizers that are adopted from GAMA~\citep{Gijsbers2021}. CLAMS aims to identify the optimal pipeline that combines the most suitable preprocessing techniques and learning algorithms. CLAMS is one of the few tools which allow full pipeline selection on clustering algorithms. CLAMS supports both internal and external metrics (if labels are provided).

\paragraph{Search Space}
By default, CLAMS contains all the scikit-learn based clustering algorithms with their hyperparameters as well as preprocessors for data cleaning and feature extraction. To the best of our knowledge, this is the first time preprocessing steps are included in the clustering pipeline in the context of automated clustering (see Table~\ref{tab:overview}). For this work, we have edited the CLAMS search space to account for the limited time during meta-training and the crashes we faced during the evaluation of multiple datasets. 
Affinity Propagation~\citep{frey2007clustering} is excluded from the search space due to its high computational time complexity, We removed spectral clustering~\citep{von2007tutorial} because of crashes on multiple datasets. Therefore, 7 clustering algorithms namely, k-Means~\citep{macqueen1967classification}, Agglomerative Clustering~\citep{gower1969minimum}, OPTICS~\citep{ankerst1999optics}, MiniBatchKMeans~\citep{sculley2010web}, DBSCAN~\citep{10.5555/3001460.3001507}, Mean Shift~\citep{1000236} and BIRCH~\citep{zhang1996birch} are selected. The detailed search space is presented in Table~\ref{tab: searchspace} in Appendix~\ref{app:search-space}.

\paragraph{Search Phase}
CLAMS includes random search (RS)~\citep{bergstra2012random}, asynchronous successive halving (ASHA)~\citep{li2020system}, and asynchronous evolutionary algorithm (ASEA) from the GAMA library for the optimization of the pipelines, based on the evaluation metric that is provided. In this work, we use the evolutionary algorithm to populate the meta-dataset. The pseudocode for the CLAMS search phase can be found in Algorithm~\ref{algo: LOTUS train}.


\subsection{Meta learning for Clustering via Dataset Similarity (CLAMS-OT)}
\begin{figure}
    \centering
    \includegraphics[width=\textwidth]{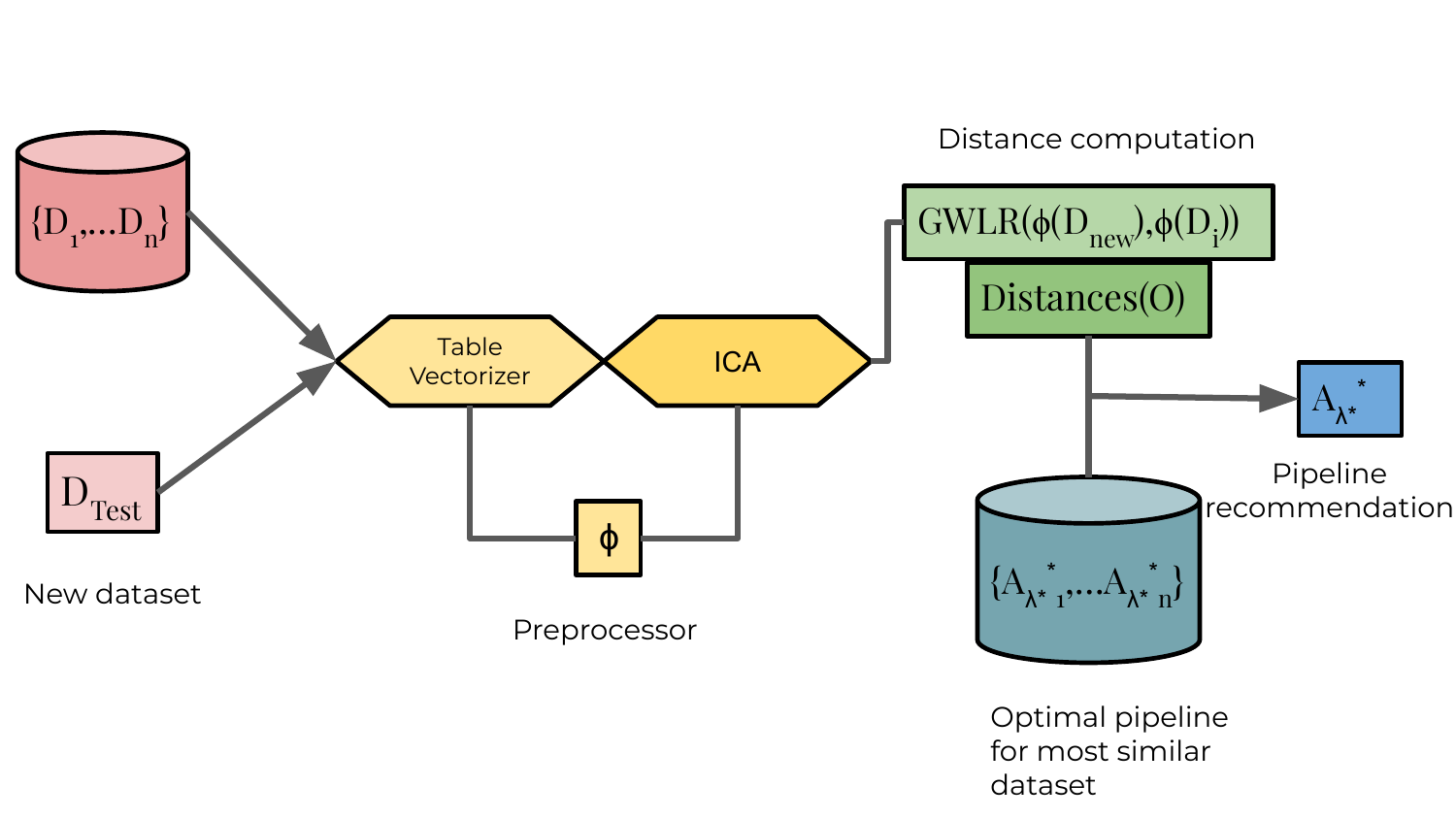}
    \caption{Overview of CLAMS-OT}
    \label{fig:clamsot}
\end{figure}

Our intuition behind the CLAMS-OT method is "If dataset A is similar to dataset B then the optimal algorithm on dataset A should perform well on dataset B". As there is no ground truth with similarity, we propose that datasets with the least distance between them from a set of datasets should be considered similar to each other. We use optimal transport distance or cost to indicate this distance between datasets. This component is visually described in Figure~\ref{fig:clamsot}.
To find the most similar dataset to the test dataset $D_{test}$ we first process the dataset via a preprocessing or transformation function $\phi$. We apply $\phi$ to every dataset in our meta-dataset $\mathcal{D}_{meta}$. The dataset distance $\mathcal{O}$ between $D_{test}$ and a single dataset from meta dataset $D_{train}$ such that $D_{train}\in\mathcal{D}_{meta}$, can be written as:

\begin{equation}
    \mathcal{O}= GW(\phi(D_{train}),\phi(D_{test}))
    \label{eq: 3-1}
\end{equation}

As we are interested in the Low-Rank Gromov-Wasserstein~\citep{Scetbon2022LowrankOT} approximation for computational efficiency from Equation \ref{eq 2-2} between these datasets, as summarized in Equation \ref{eq: 3-2}, where $r$ is the selected rank, This distance is described as.

\begin{equation}
    \mathcal{O}=  \text{GW-LR}^{(r)}(\phi(D_{train}),\phi(D_{test}))
\label{eq: 3-2}
\end{equation}
The most similar prior dataset $D_{similar} \in \mathcal{D}_{meta}$ is the dataset with the smallest distance to the new dataset $D_{test}$ from a set of pairwise distances between datasets. CLAMS-OT then suggests $A^*_{\lambda^{*}_{similar}} \in \mathcal{A}$ as the optimal configuration for $D_{test}$, as shown in Algorithm \ref{algo: LOTUS train}.

\begin{algorithm}[tb]
    \caption{Pseudocode for CLAMS (Meta-training)}\label{algo: LOTUS train}
    \begin{flushleft}
        \textbf{Inputs:} $\mathcal{D}_{meta}, L, \boldsymbol{A},
        \Lambda_A$ for each $A \in \boldsymbol{A}$.
    \end{flushleft}
    \begin{algorithmic}[1]
            \FOR{$D_{i} \in \mathcal{D}_{meta}$}
            \STATE $A^*_{\lambda^*_i} \gets \operatorname*{argmin}_{%
       \substack{%
         A \in \boldsymbol{A} \\
         \lambda \in \Lambda_A
       }
     }
 L \left( A_\lambda, D_i \right)
     $
         \STATE Add $A^*_{\lambda^*_i}$ to $\mathcal{A}$.

            \ENDFOR
    \end{algorithmic}
\end{algorithm}

\begin{algorithm}[tb]
    \caption{Pseudocode for CLAMS-OT (meta-testing)}\label{algo: LOTUS}
    \begin{flushleft} 
    \textbf{Inputs:} $D_{test}, \mathcal{D}_{meta}, \mathcal{A}$
    \end{flushleft}
\begin{algorithmic}[1]
    \FOR{$D_{i} \in \mathcal{D}_{meta}$}
    \STATE $\mathcal{O}_i \gets GWLR(\phi(D_{test}), \phi(D_i))$    \COMMENT{Distance calculation}
    \ENDFOR
    \STATE $s \gets {\mathrm{argmin}}\{\mathcal{O}_1,...,\mathcal{O}_n\}$    \COMMENT{Retrieval of most similar dataset}
    \STATE $A^*_{\lambda^{*}_{new}} \gets A^*_{\lambda^{*}_s}$    \COMMENT{Model Selection}
    \end{algorithmic}
\end{algorithm}
\section{Experimental Setup}
For our experiments, we use 57 datasets from OpenML~\citep{OpenML2013} which are suitable for clustering. The full list of datasets and their dimensionalities are shown in Table~\ref{tab:omldata} in Appendix~\ref{app:datasets}. These datasets are selected manually, we aim to select both synthetic and real-world datasets for our setting. 
\par
We cannot directly perform distance calculations on datasets with missing or non-numeric data, so we employ a preprocessing pipeline before computing distances.
First, we convert our non-numeric data in our datasets using the TableVectorizer in the dirty-cat library~\citep{dirtycat, dirtycat2}. The TableVectorizer encodes nominal values using one-hot encoding for low cardinality features or GaP encoding~\citep{10.1145/1008992.1009016} for high cardinality features.
Next, we transform the datasets via Independent Component Analysis~\citep{Hyvrinen2000IndependentCA}, using scikit-learn's implementation, as we found this speeds up distance computation considerably and improves its robustness. 
Finally, we convert them into JAX pointclouds geometry objects\footnote{\url{https://ott-jax.readthedocs.io/en/latest/\_autosummary/ott.geometry.pointcloud.PointCloud.html}} and define a quadratic regularized optimal transport problem. Using the Gromov Wasserstein Low-Rank solver from OTT-JAX~\citep{Cuturi2022OptimalTT} for this quadratic problem, we obtain the distance (cost) between the two pointclouds.\footnote{We set the rank parameter of GWLR to \textit{6}, which we found results in a good accuracy-speed tradeoff.} When given a new dataset, the pipeline corresponding to the dataset with the lowest distance (excluding the new dataset itself) is chosen from the optimal pipeline database, as illustrated in Figure \ref{fig:clamsot}.

\section{Results}
To evaluate the effectiveness of our approach, we report the average Adjusted Mutual Information (AMI)~\citep{JMLR:metrics} obtained from five runs of both the baselines and our approach (CLAMS-OT). We utilize the Bayesian Wilcoxon signed-rank test, or the region of practical equivalence (ROPE) test~\citep{Ropetutorial,rope2}, to analyze the experimental results. The ROPE test defines an interval within which differences in model performance are considered equivalent to each other. It returns probabilities based on the measured performance. One model is considered to be better than the other based on the region of practical importance. By setting the ROPE value to 1\%, we can make more practical comparisons between model performances. To run and visualize the analysis, we use the baycomp library~\citep{Ropetutorial}. Our experimental results, as determined by the ROPE test, demonstrate the superiority of our approach over the baselines. 

We compare our approach to default configurations of many different clustering algorithms and show the results in Figure \ref{rope}. 
\\
\begingroup
\begin{figure}
\centering
\setlength{\unitlength}{2.5cm}
\begin{picture}(6,3)
\put(0,0){\includegraphics[width=6cm]{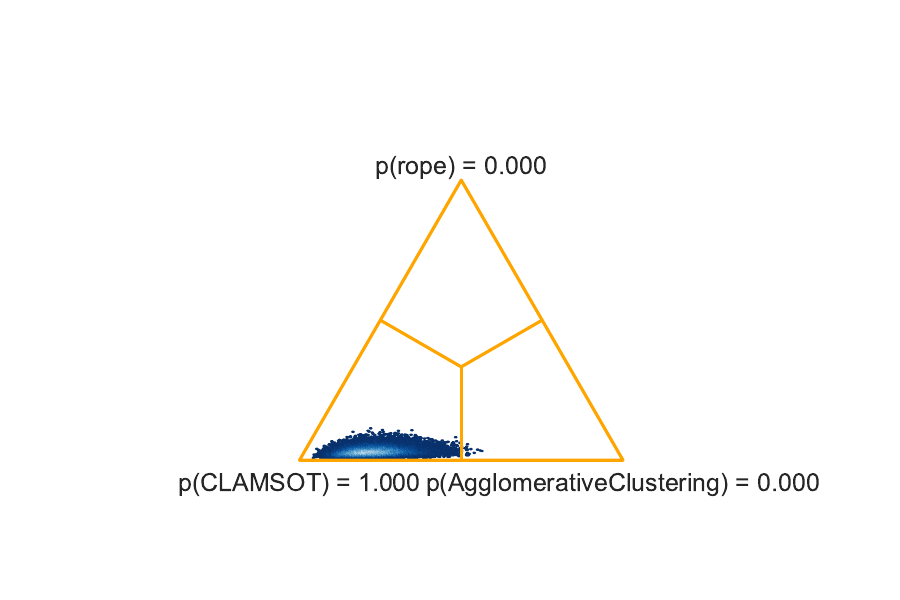}}
\put(2,0){\includegraphics[width=6cm]{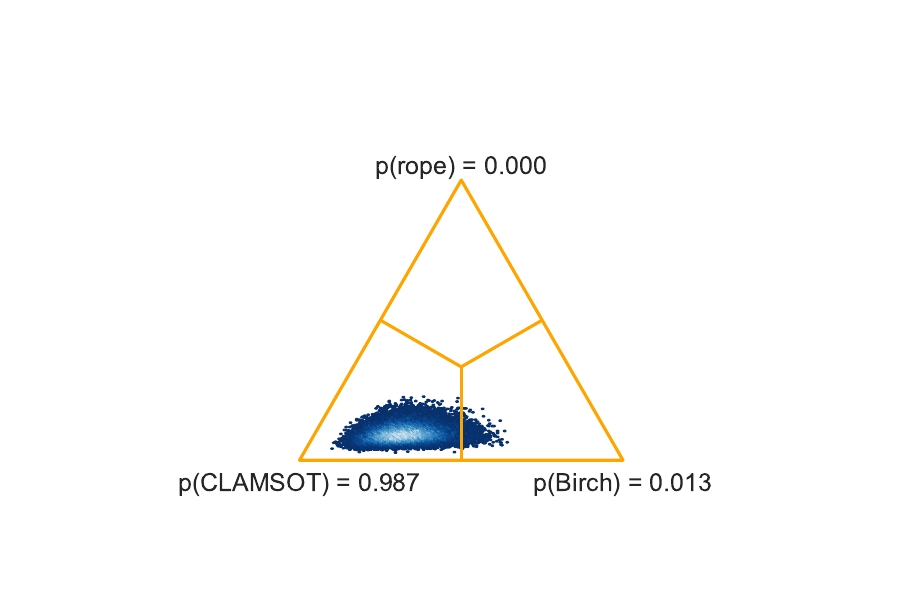}}
\put(4,0){\includegraphics[width=6cm]{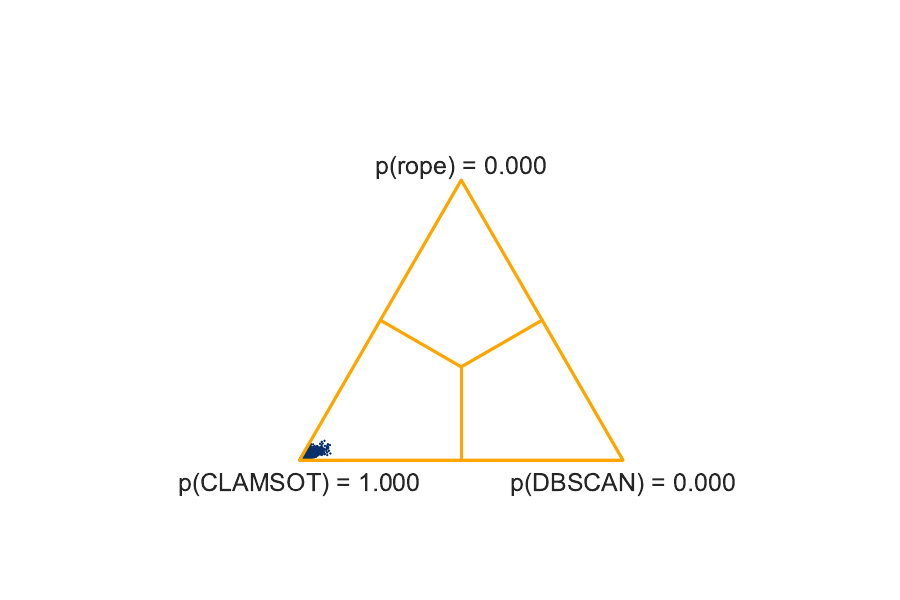}}
\put(0,1){\includegraphics[width=6cm]{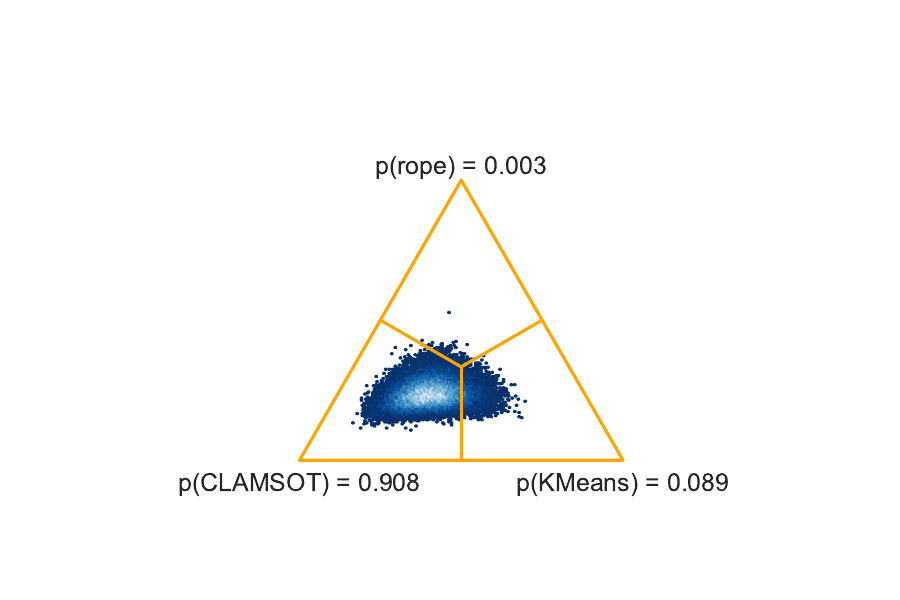}}
\put(2,1){\includegraphics[width=6cm]{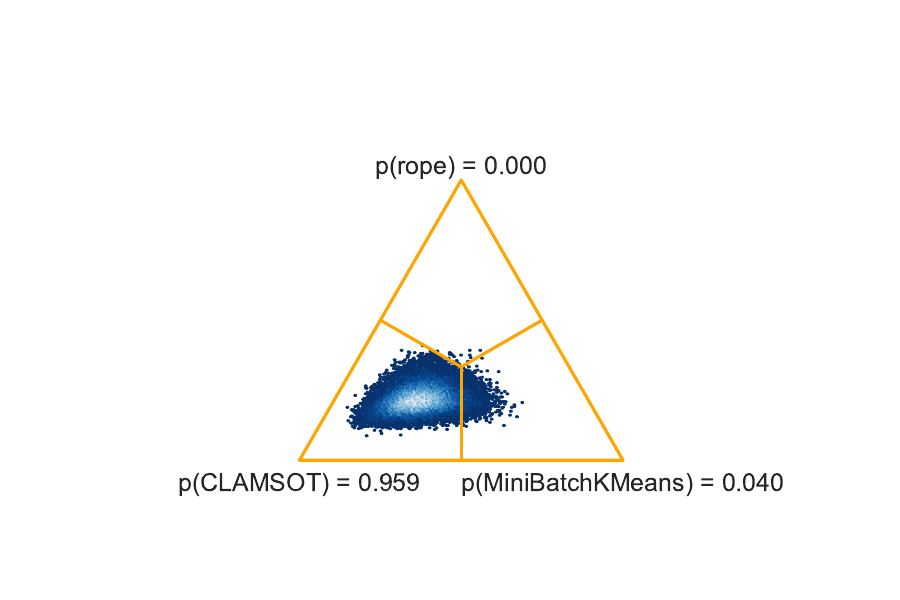}}
\put(4,1){\includegraphics[width=6cm]{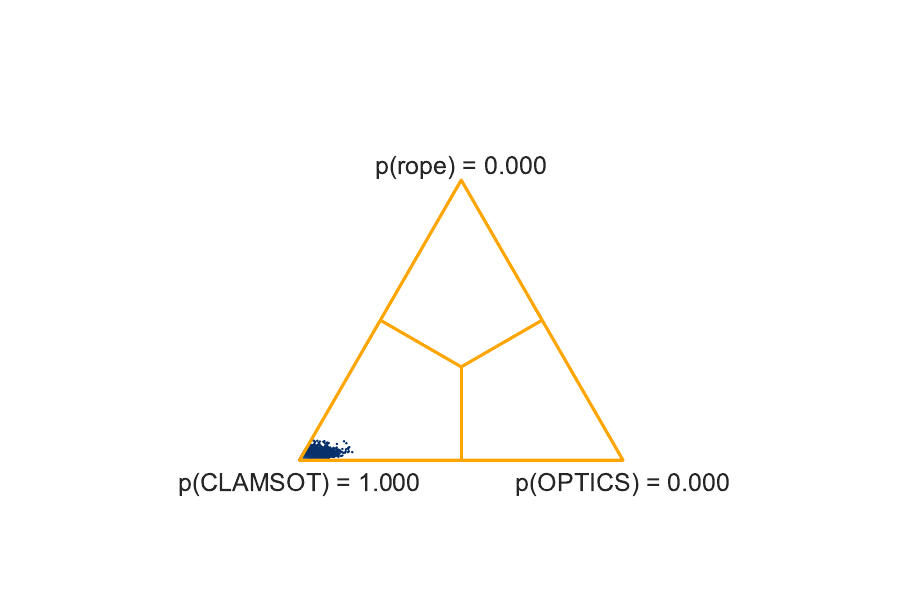}}
\end{picture}
\caption{Bayesian Wilcoxon signed-rank test results of CLAMS-OT vs Baselines with ROPE=0.01, this figure shows the simplex and projections of the posterior for the Bayesian sign-rank test.
 The closer the distribution is to the bottom left corner, the more likely it is that our method is better.}
\label{rope}
\end{figure}
\endgroup

\begin{figure}
    \centering
    \includegraphics[width=\textwidth]{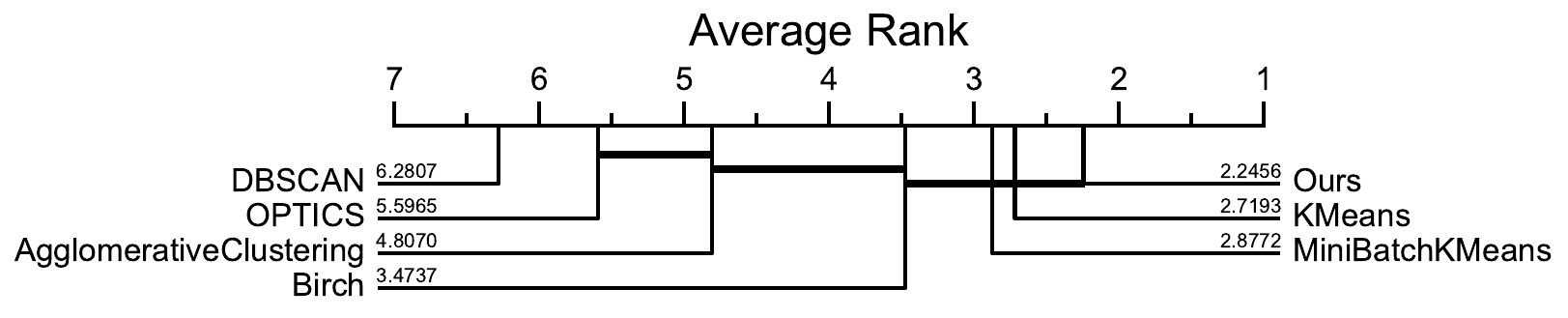}
    \caption{Critical difference diagram of CLAMS-OT vs baselines (Mean Adjusted mutual information score)}
    \label{fig:cd}
\end{figure}

We provide a critical difference diagram\footnote{ We use the code provided here \url{https://github.com/hfawaz/cd-diagram } to generate the critical difference diagram for our experiment} in Figure \ref{fig:cd}. Our approach also gives the lowest rank among all the candidates. 
\section{Discussion}

\subsection{OT as an Indicator of Dataset Similarity}
\begin{figure}
     \centering
     \includegraphics[width=10cm]{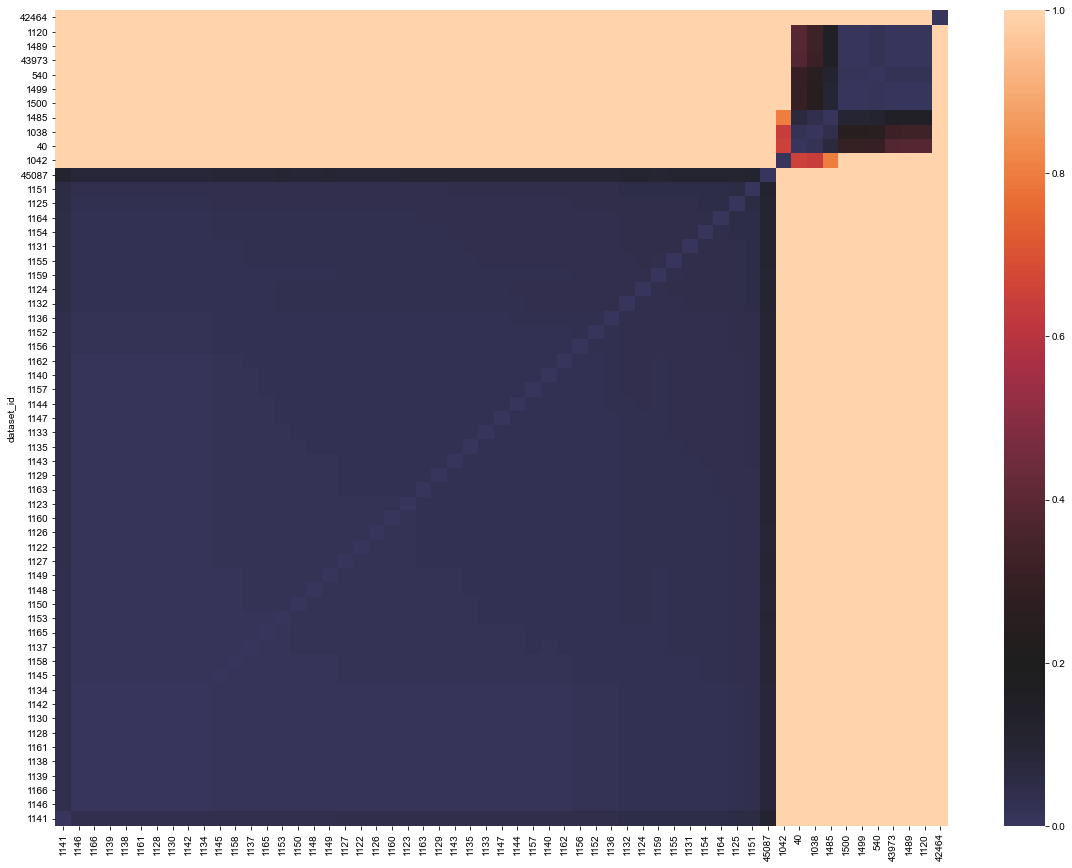}
     \caption{Heatmap showing dataset similarity,  lesser value means more similar}
     \label{fig: heatmap}
 \end{figure}

 Our research on the dataset and task similarity builds on existing studies that utilize distance measures to quantify the similarity between datasets. Specifically, we draw inspiration from the work of \cite{AlvarezMelis2020GeometricDD}, who proposed the Optimal Transport Dataset Distance (OTDD), a similarity metric between datasets that utilizes optimal transport to learn a mapping over the joint feature and label spaces. There have been other studies exploring the space of dataset and task similarity with distance measures. \cite{pmlr-v139-gao21a} propose ``coupled transfer distance'' which utilizes optimal transport distances as a transfer learning distance metric on image data.~\citet{10.1093/imaiai/iaaa033} explore connections between Deep Learning, Complexity Theory, and Information Theory through their proposed asymmetric distance on tasks. In this work, we explored Entropic Gromov Wasserstein distance as a means to signal dataset similarity in a data suite. Our study presents a heatmap of GWLR distances of every dataset in $\mathcal{D}_{meta}$ in Figure \ref{fig: heatmap}, highlighting the differences between various datasets. We exclude very high distances from this figure to show the finer details of dataset similarity.
 \subsection{Limitations}
Our framework has limitations that should be taken into consideration. First, CLMASOT effectiveness depends on the presence of similar datasets to the meta-test dataset within the $\mathcal{D}{meta}$. In cases where there are no similar datasets, such as with dataset id \textit{42464} in our experiments, our suggested pipeline may not yield favorable results. Second, the time complexity of our system scales linearly with the number of datasets in the $\mathcal{D}{meta}$. This means that as the number of datasets increases, CLAMS-OT may require more time to perform model selection. These limitations are important to note as they may impact the practical application of our approach in certain scenarios.
\subsection{Future Work}
This work proposed a system to automate clustering with zero-shot recommendation via optimal transport distance. Optimal transport distance is still very expensive to compute. Though the low-rank computation decreases the time complexity, the system still takes around 30 minutes to compute similarity with other datasets. We believe this can be largely improved by using Wasserstein Embedding Networks~\citep{courty2018learning} or MetaICNN~\citep{Amos2022MetaOT} for faster computation, though these networks still do not support Gromov Wasserstein space.  Another direction where we believe this system can be optimized is to use dynamic training time depending on datasets during the CLAMS search phase.

\section{Conclusion}
In this work, we have presented CLAMS, An AutoML system for selecting clustering algorithms, and CLAMS-OT, a zero-shot model recommendation system to recommend clustering models for datasets without labels. CLAMS allows performing automated clustering on a given dataset via a given CVI, CLAMS-OT is a meta-learner that uses optimal transport distances. Our system achieved the highest performance in the ROPE test and the lowest rank in the critical difference test. We discussed the motivation and analysis of our meta-learning technique.

\bibliography{biblo.bib}


\clearpage
\newpage

\appendix
\section{Search Space.}
\label{app:search-space}

\begin{table}[h]
\begin{center}
\begin{minipage}{\textwidth}
\caption{Clustering algorithms and hyperparameter space}
\label{tab: searchspace}%
\begin{tabular}{@{}lll@{}}
\toprule
Algorithm & Hyperparameter  & Search Space\\
\midrule
k-Means    & n\_clusters   & [2-21]   \\
           & n\_init\   & ['auto']      \\
           & max-iter   & [300-500]      \\
           & algorithm & ['lloyd', 'elkan'] \\
MiniBatchKmeans  & n\_clusters   & [2, 21]   \\
                & max-iter   & [100-500]      \\
                & min-bin-freq & [1,2,3,4,5]\\
Mean Shift & bin\_seeding & [True, False] \\
           & min\_bin\_freq & [1-5]) \\
           & max\_iter & [2-300]      \\
AgglomerativeClustering & n\_clusters & n [2-21]   \\
           & affinity & [‘euclidean’, ‘manhattan’, ‘cosine’, ‘l1’, ‘l2’] \\
           & linkage & [‘ward’, ‘complete’, ‘average’, ‘single’] \\
DBSCAN & eps & [0.1-0.5]\\
       & min\_samples & [3,4,5,6,7,8] \\
       & p & [1, 2] \\
OPTICS & min\_samples & [3,4,5,6,7,8\\
       & p & [1, 2] \\
       & xi & [0.05-5] \\
BIRCH & n\_clusters   &  [2-21]   \\
      & threshold     & [0.2-0.8] \\
      & branching\_factor & [25, 50, 75] \\\hline
        
\end{tabular}
\end{minipage}
\end{center}
\end{table}

\section{Datasets}
\label{app:datasets}
\begin{longtable}{|c|c|c|c|c|} 
\hline
\textbf{id} & \textbf{dataset name}     & \textbf{number of instances} & \textbf{number of features} & \textbf{number of classes} \\ \hline
40          & sonar                     & 208                          & 61                          & 2                          \\
540         & mu284                     & 284                          & 10                          & 50                         \\
1038        & gina\_agnostic            & 3468                         & 971                         & 2                          \\
1042        & gina\_prior               & 3468                         & 785                         & 2                          \\
1120        & MagicTelescope            & 19020                        & 12                          & 2                          \\
1122        & AP\_Breast\_Prostate      & 413                          & 10936                       & 2                          \\
1123        & AP\_Endometrium\_Breast   & 405                          & 10936                       & 2                          \\
1124        & AP\_Omentum\_Uterus       & 201                          & 10936                       & 2                          \\
1125        & AP\_Omentum\_Prostate     & 146                          & 10936                       & 2                          \\
1126        & AP\_Colon\_Lung           & 412                          & 10936                       & 2                          \\
1127        & AP\_Breast\_Omentum       & 421                          & 10936                       & 2                          \\
1128        & OVA\_Breast               & 1545                         & 10936                       & 2                          \\
1129        & AP\_Uterus\_Kidney        & 384                          & 10936                       & 2                          \\
1130        & OVA\_Lung                 & 1545                         & 10936                       & 2                          \\
1131        & AP\_Prostate\_Uterus      & 193                          & 10936                       & 2                          \\
1132        & AP\_Omentum\_Lung         & 203                          & 10936                       & 2                          \\
1133        & AP\_Endometrium\_Colon    & 347                          & 10936                       & 2                          \\
1134        & OVA\_Kidney               & 1545                         & 10936                       & 2                          \\
1135        & AP\_Colon\_Prostate       & 255                          & 10936                       & 2                          \\
1136        & AP\_Lung\_Uterus          & 250                          & 10936                       & 2                          \\
1137        & AP\_Colon\_Kidney         & 546                          & 10936                       & 2                          \\
1138        & OVA\_Uterus               & 1545                         & 10936                       & 2                          \\
1139        & OVA\_Omentum              & 1545                         & 10936                       & 2                          \\
1140        & AP\_Ovary\_Lung           & 324                          & 10936                       & 2                          \\
1141        & AP\_Endometrium\_Prostate & 130                          & 10936                       & 2                          \\
1142        & OVA\_Endometrium          & 1545                         & 10936                       & 2                          \\
1143        & AP\_Colon\_Omentum        & 363                          & 10936                       & 2                          \\
1144        & AP\_Prostate\_Kidney      & 329                          & 10936                       & 2                          \\
1145        & AP\_Breast\_Colon         & 630                          & 10936                       & 2                          \\
1146        & OVA\_Prostate             & 1545                         & 10937                       & 2                          \\
1147        & AP\_Omentum\_Kidney       & 337                          & 10936                       & 2                          \\
1148        & AP\_Breast\_Uterus        & 468                          & 10937                       & 2                          \\
1149        & AP\_Ovary\_Kidney         & 458                          & 10936                       & 2                          \\
1150        & AP\_Breast\_Lung          & 470                          & 10936                       & 2                          \\
1151        & AP\_Endometrium\_Omentum  & 138                          & 10936                       & 2                          \\
1152        & AP\_Prostate\_Ovary       & 267                          & 10936                       & 2                          \\
1153        & AP\_Colon\_Ovary          & 484                          & 10936                       & 2                          \\
1154        & AP\_Endometrium\_Lung     & 187                          & 10936                       & 2                          \\
1155        & AP\_Prostate\_Lung        & 195                          & 10936                       & 2                          \\
1156        & AP\_Omentum\_Ovary        & 275                          & 10936                       & 2                          \\
1157        & AP\_Endometrium\_Kidney   & 321                          & 10936                       & 2                          \\
1158        & AP\_Breast\_Kidney        & 604                          & 10936                       & 2                          \\
1159        & AP\_Endometrium\_Ovary    & 259                          & 10936                       & 2                          \\
1160        & AP\_Colon\_Uterus         & 410                          & 10936                       & 2                          \\
1161        & OVA\_Colon                & 1545                         & 10936                       & 2                          \\
1162        & AP\_Ovary\_Uterus         & 322                          & 10936                       & 2                          \\
1163        & AP\_Lung\_Kidney          & 386                          & 10936                       & 2                          \\
1164        & AP\_Endometrium\_Uterus   & 185                          & 10936                       & 2                          \\
1165        & AP\_Breast\_Ovary         & 542                          & 10936                       & 2                          \\
1166        & OVA\_Ovary                & 1545                         & 10936                       & 2                          \\
1485        & madelon                   & 2600                         & 501                         & 2                          \\

1499        & seeds                     & 210                          & 8                           & 3                          \\
1500        & seismic-bumps             & 210                          & 8                           & 3                          \\
42464       & Waterstress               & 1188                         & 23                          & 2                          \\
43973       & phoneme                   & 3172                         & 6                           & 2                          \\
45087       & Colon                     & 6                            & 2001                        & 2                          \\ \hline
\caption{Table of OpenML datasets used} 
\label{tab:omldata}
\end{longtable}
\end{document}